\documentclass[11pt,a4paper]{article}
\usepackage[hyperref]{emnlp2020}
\usepackage{times}
\usepackage{latexsym}

\usepackage{microtype}

\aclfinalcopy 

\usepackage{graphicx}
\usepackage{caption}
\usepackage{subcaption}
\usepackage{hyperref}

\title{Six Attributes of Unhealthy Conversations}

\author{
Ilan Price \\
  University of Oxford \\
  ilan.price@maths.ox.ac.uk \\ \And
  Jordan Gifford-Moore \\
  jordan.gifford-moore@flinders.edu.au \\ \AND
  Jory Flemming \\
  University of South Carolina \\
  fleminj6@mailbox.sc.edu\\\ \And
  Saul Musker \\
  saul@presidency.gov.za \\\And
  Maayan Roichman \\
  University of Oxford \\
  maayan.roichman@anthro.ox.ac.uk\\\AND
  Guillaume Sylvain \\
  gs1867@nyu.edu\\\And
  Nithum Thain \\
  Google Brain \\
  nthain@google.com\\\AND
  Lucas Dixon \\
  Google Research \\
  ldixon@google.com\\\And
  Jeffrey Sorensen\\
  Jigsaw\\
  sorenj@google.com
  }

\date{}

\begin{document}
\maketitle
\begin{abstract}
We present a new dataset of approximately $44000$ comments labeled by crowdworkers. Each comment is labelled as either `healthy' or `unhealthy', in addition to binary labels for the presence of six potentially `unhealthy' sub-attributes: (1) hostile; (2) antagonistic, insulting, provocative or trolling; (3) dismissive; (4) condescending or patronising; (5) sarcastic; and/or (6) an unfair generalisation. Each label also has an associated confidence score.  We  argue that there is a need for datasets which enable research based on a broad notion of `unhealthy online conversation'. We build this typology to encompass a substantial proportion of the individual comments which contribute to unhealthy online conversation. For some of these attributes, this is the first publicly available dataset of this scale.  We explore the quality of the dataset, present some summary statistics and initial models to illustrate the utility of this data, and highlight limitations and directions for further research.

\end{abstract}

\section{Introduction}

Analysis of online user discussion continues to be a critical area of interdisciplinary research. Increasing rates of internet access and the development of a diverse range of online forums has allowed for conversation between individuals across the globe on an extraordinary range of topics. However, this has been accompanied by a surge in abuse and other negative behaviours online, the impacts of which have been well-documented in academic research. It has been found that targeted negative comments and harassment online can seriously impact individual well-being \cite{weingartner2019online, bauman2013cyberbullying}, force users to leave a community or reduce online participation \cite{wulczyn2017ex, blackburn2014stfu}, and potentially lead to offline hate-crimes  \cite{mulki2019hsab,hassan2018exposure}. While these forms of comments may be explicit or overtly harmful, they are also often difficult to detect or ambiguous. Where there are insufficient moderation resources to scale with a forum’s user-base, this can lead to unchecked negative discourse, or cause website administrators to restrict user comment functions. This means that research which aims to enable automated moderation, provide a review triage service for human moderation teams, or design systems to nudge users towards healthier conversation, has significant potential for contributing to both the availability and quality of online discourse. 

A persistent challenge for researchers and site administrators in this area is the need to: (a) establish a typology of comments which are undesirable in online discussions; (b) apply this typology in a consistent and reliable manner; and (c) account for adversarial user behaviour in response to moderation. This is complicated by the fact that there is no single objective set of categories for speech which ought to be excluded in all contexts, with perceptions of undesirable speech differing across individuals, cultures, geographies, and online communities \cite{vidgen2019challenges}. 

Prior research on toxic comments online has found that classifiers trained on crowdsourced data can be effective at detecting the most overt forms of toxic comments. However, there remain difficulties in detecting subtler forms of toxicity which may be implicit, require idiosyncratic knowledge, familiarity with the conversation context, or familiarity with particular cultural tropes \cite{kohlipaying, van2018challenges, parekh2017toxic}. One of the key ingredients to progress on this front will be high quality, large, annotated datasets addressing these more subtle harmful attributes, from which machine learning models will be able to learn. Unfortunately, for most subtler toxic attributes there are few available datasets (or none, particularly in many languages other than English), which is a bottleneck preventing further research \cite{fortuna2019hierarchically}.

We aim to contribute to research in this area through the release of the Unhealthy Comment Corpus (UCC) of approximately 44,000 comments and corresponding crowdsourced labels and confidence scores. The labelling typology for the dataset identifies for each comment a higher-level classification of whether that comment `has a place in a healthy online conversation', accompanied for each comment by binary labels for whether it is: (1) hostile, (2) antagonistic, insulting, provocative or trolling (together, `antagonistic'),  (3) dismissive, (4) condescending or patronising (together, `condescending'), (5) sarcastic, and/or (6) an unfair generalisation. For each label there is also an associated confidence score (between 0.5 and 1). The UCC is open source and available on Github.\footnote{\href{https://github.com/conversationai/unhealthy-conversations}{github.com/conversationai/unhealthy-conversations}} 

The UCC contributes further high quality data on attributes like sarcasm, hostility, and condescension, adding to existing datasets on these and related attributes \cite{wang2019talkdown, davidson2017automated, wulczyn2017ex, chen2017presenting}, and provides (to the best of our knowledge) the first dataset of this scale with labels for dismissiveness, unfair generalisations, antagonistic behavior, and overall assessments of whether those comments fall within `healthy' conversation. We also make use of and illustrate the benefits of annotator trustworthiness scores when crowdsourcing labels on subjective data of this sort. 

This paper is  structured as follows. Section \ref{sec: background} outlines the motivation and background to the UCC attribute typology. Section \ref{sec: data collection} details the data collection and quality control processes. In Section \ref{sec: dataset} we present some summary statistics, benefits, and limitations of  the data, and in Section \ref{sec:  model} we present a baseline classification model for this dataset, and evaluate  its performance. Section \ref{sec: biases} highlights potential sources of bias in this dataset, and the need to be cognisant of these when conducting further research in this area \cite{dixon2018measuring}.

\section{From `toxic' comments to `unhealthy' conversation}\label{sec: background}

In this paper, we broadly characterise a healthy online public conversation as one where posts and comments are made in good faith, are not overly hostile or destructive, and generally invite engagement. Such a conversation may include robust engagement and debate, and is generally (though not always) focused on substance and ideas. Importantly, though, healthy contributions to online conversations are not necessarily friendly, grammatically correct, well constructed, intellectual, substantive, or even free of any vulgarity.

Some harmful contributions to conversations are obviously derogatory, threatening, violent, or insulting \cite{anderson2018toxic}, and these are the sorts of comments which have been the primary focus of research in algorithmic moderation assistance and related areas. However, many of those comments which deter people from engagement or create downward spirals in interactions can be more subtle \cite{zhang2018conversations}. This is especially the case with conversations \textit{online}, many of which (i) take place in a `public’ forum that is visible to thousands of others, and (ii) involve strangers who have never met and know little about one another \cite{santana2014virtuous}. These two features of online conversations can sometimes enhance commenters' sensitivity to subtler forms of toxicity like sarcasm, condescension, or dismissiveness, amplifying their negative impact on conversations despite the fact that these attributes may be less (or not at all) harmful in other specific contexts. 

Identifying subtle indicators of problematic online comments is a difficult task. There are at least three reasons for this. First, they are less extreme and therefore less likely to use clearly identifiable explicit or inflammatory language. Second, a substantive point might be made in an inflammatory way, or a remark may be perceived differently depending on the context, norms, and expectations of the reader. Third, there is an even greater risk of identifying `false positives' and `false negatives', since many of the expressions used in subtle forms of toxicity can also be deployed for positive contributions. For example, sarcasm is often used in derisive or bullying ways, but it can also be used for humour or to express a substantive, inoffensive point \cite{vidgen2019challenges}.

The challenge is to identify the subtle characteristics of harmful comments online despite their ambiguity, without falsely identifying healthy comments. We differentiate between two categories. The first, which is the most well studied to date, are those whose explicit intention is to insult, threaten, or abuse. The second category, are comments which engage with others, share an opinion, or contribute to the conversation, but are written in a way which is likely to antagonise, hurt, or deter others. We  found these comments to be at least as prevalent in the sample data (Table \ref{table: proportions}). Our typology of unhealthy attributes aims to include this second category of comments, and determine whether annotators believe they belong in a healthy online conversation.

\begin{figure}[!h]
    \hspace{0.1cm}
    \includegraphics[width = 0.48\textwidth]{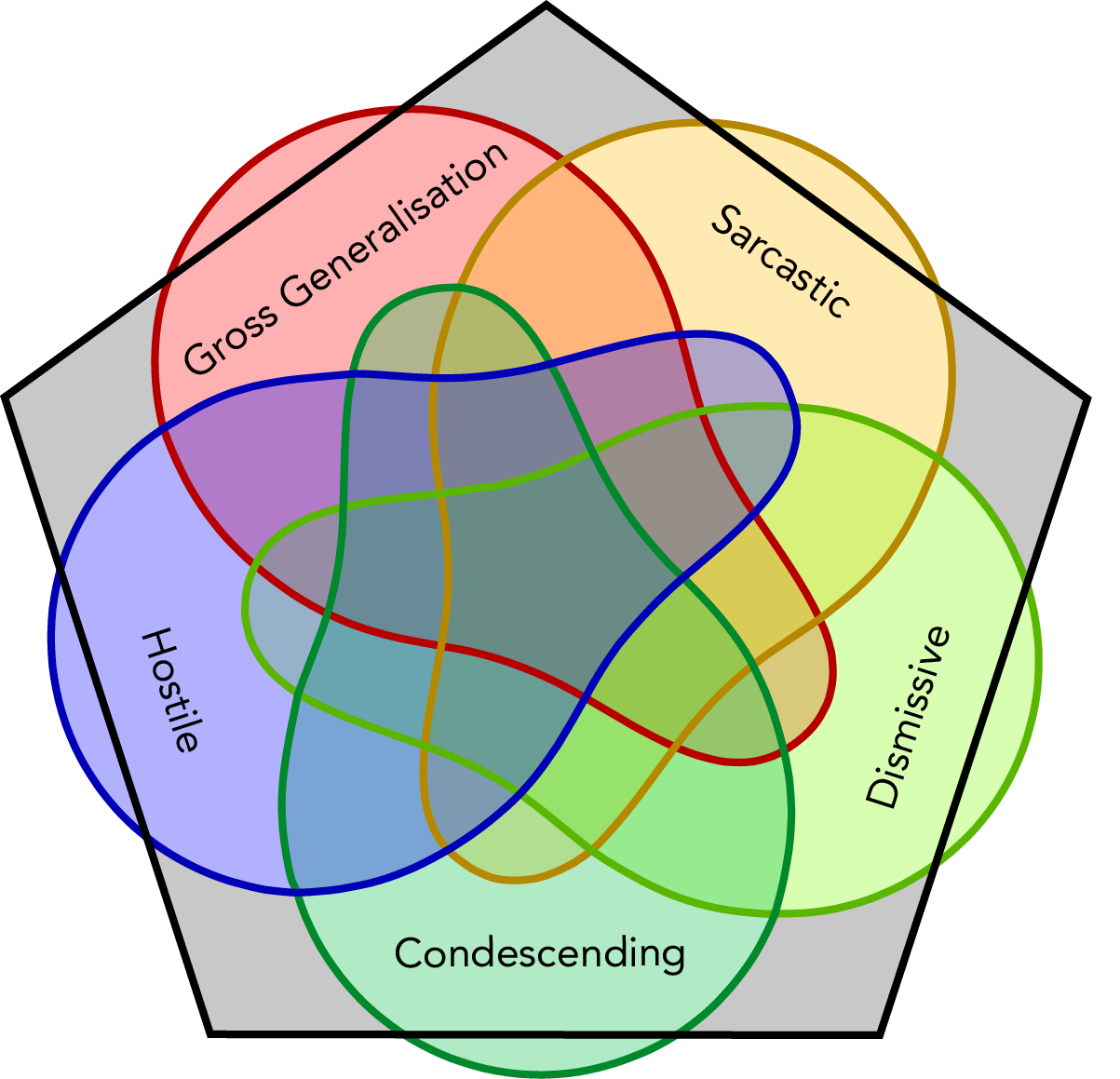}
    \caption{A visualisation of the proposed typology of  unhealthy online comments. The grey pentagon represents unhealthy comments. Note that in this figure, `hostile' and `antagonistic' are represented jointly as `hostile'.}
    \label{fig:attribute venn}
\end{figure}

Our hypothesis was that together these 6 attributes account for the majority of `unhealthy' comments online, but that there will still be some comments that are `unhealthy' but do not display any sub-attribute, and also some which are `healthy' despite representing one or more sub-attributes (see Figure \ref{fig:attribute venn}). In general, whether the presence of these attributes indicates healthy or unhealthy conversation will also depend significantly on the nature of the forum and users. Nonetheless, the combination of an abstract `health' rating with the other 6 attributes provides a useful dataset for investigating nuanced comments, and could be used to help develop a broader range of models that are customised for specific production environments. 
 
\section{Source data and annotation}\label{sec: data collection}
The dataset comprises randomly chosen comments from the Globe and Mail news site (sampled from the SFU Opinion and Comment Corpus dataset) \cite{kolhatkar2019sfu}, of 250 characters or less. Comment scores were crowdsourced using Figure Eight (now Appen). The annotation job consisted of 588 crowdworkers (annotators) providing 244468 judgements on 44355 comments.\footnote{According to statistics provided by Appen, the average time spent on those annotations which were included in the final dataset was between 12 and 13 seconds per comment.} Each annotator was asked to identify for each comment whether it was healthy and if any of the attributes were present, in the form of a standard questionnaire (see Appendix \ref{app: questionnaire}). Annotators were not given any wider context or additional information about where a comment was posted or how it was engaged with by other users.

To both accommodate and attempt to resolve meaningful disagreement, we applied a dynamic judgement method which requests additional annotations for those comments on which there was insufficient consensus (either yes or no with a confidence of less than 75\%). All comments were annotated at least three times, and more annotators were added, up to a limit of five annotators per comment until sufficient consensus was reached.

\noindent \textbf{Annotation Job Refinement.}
The inherent subtlety, subjectivity, and frequent ambiguity of the attributes covered in this dataset make crowdsourcing quality attribute labels an unavoidably difficult process.

Typically the goal in an annotation task would simply be to maximise agreement between the multiple annotators of each comment. However, when the annotation task is inherently subjective and meaningful difference of opinion is itself valuable data, the goal becomes instead to maximise common understanding of the task across annotators. This entails tailoring the phrasing of the questions put to annotators, so as to create as common an understanding as possible of what each question is really asking. This way, disagreement between annotators reflected in the dataset will represent different reasonable readings of the same comment which are themselves important to capture. In research on irony and sarcasm, for example, Filatova noted the difficulty even among expert researchers in formally defining these terms \cite{filatova2012irony}. For the other attributes included in this dataset which are as (if not more) ambiguous and subtle than sarcasm, we expect this to hold true as well.

The exact wording of each question on the questionnaire went through multiple iterations, tested by smaller scale experiments to evaluate  effectiveness. The quality of the resulting data was evaluated manually by our team, calculating the proportion of perceived mistaken annotations and their `severity': to what extent a judgement was `obviously wrong', as opposed to an understandable alternative reading of a comment. 

We found that providing annotators with precise and more comprehensive definitions of each attribute was not more likely to produce inter-annotator agreement or better quality data. Neither, however, were best results produced by asking simple, `yes or no' questions such as `Is this comment dismissive?' for all attributes. The best results were achieved by relying primarily on annotators implicit understandings of and intuitions about the attributes, aided by brief inline explanations. We added explanations to avoid mistakes for those attributes which are more ambiguous, and for which our smaller tests had indicated required further guidance. These can be seen in the questionnaire included as Appendix \ref{app: questionnaire}.

To ensure that disagreement reflects reasonable difference of opinion, rather than inattention or misunderstanding of the task, it is necessary to apply a method of quality control. The attempt to create a labeled dataset is premised on the assumption of some `ground truth'; that it is possible for comments to have labels and confidence scores accurately representing the presence of one or more attributes to some extent. However, the extent to which a comment displays one or more attribute is subjective, and the scores would be unhelpful if they did not capture what a wider and more diverse audience than our team of authors would understand the comments to mean. Our process of quality control therefore aimed to reduce the number of `bad' annotators, those who either do not understand or appropriately engage with the task, while still allowing for differences of opinion. 

Our primary quality control mechanism was to collate a set of `test comments', for which we had manually established the correct answers. Annotators encountered one test comment per  batch of seven comments they reviewed, without knowing which of the seven was the test comment,  and their running accuracy on these test comments was defined as their `trustworthiness score'.  The task required that annotators maintain a trustworthiness score of more than 78\%. If an annotator dropped below this level, they were removed from the annotator pool for this task, and all of their prior annotations were discarded\footnote{This was a threshold selected through initial test jobs, to balance budget and quality considerations. A higher threshold yields more trustworthy annotations, but consequently discards more existing data when annotators drop below that threshold.}. The removed `bad' annotator judgements were replaced by newly collected trusted judgements as necessary.

We restricted our test comments to what were (in our view) clear and definitive examples of the attributes, such that one would fail on the test comments only if one has an incorrect understanding of what is meant by a particular attribute. In the course of our preliminary small-scale refining iterations of the questionnaire, analysis of responses revealed some recurring misunderstandings or mistakes. For example, a  common  error was to label all non-sarcastic humour as sarcasm, or to conflate polite disagreement with dismissiveness. As a result, we identified and included specific test comments, drawn from real examples, aimed at reducing these common errors. 

We included very few test comments for the higher level question on whether a comment belongs in a healthy conversation. Any test questions on this topic were very extreme examples, such as highly abusive explicit comments, to ensure that annotators were not randomly answering that question. We had two reasons for minimising the use of test comments for this question. Firstly, since this was in our view the most open-ended question, it is difficult to establish tests on the basis of which to exclude annotators. Secondly, allowing greater annotator discretion on this question provides insight on whether there is a correlation between the six attributes and being labelled as unhealthy.\footnote{There remains a clear methodological issue with using this data for comparing the set of comments classed as `unhealthy' with those classed as one or more of the other attributes: having been asked all questions as part of the same questionnaire, annotators may have been primed to associate the attributes with `unhealthiness', even if they would not have done so otherwise. } 

\section{The UCC dataset}\label{sec: dataset}

The  dataset comprises a total  of 44355 comments labelled  `yes' or `no' for each attribute, along with a confidence score for each label. The labels and corresponding confidence scores for each attribute are based on an aggregation of the answers given by different annotators, weighted by their respective `trustworthiness' scores. As an example to demonstrate this process, consider a comment annotated by 5 annotators with trustworthiness scores 0.78, 0.85,  0.9, 1.0,  and 0.95, who judge  a comment for a particular attribute with judgements `yes', `yes', `yes', 'no', 'yes' respectively. Let $T$ be the sum of their trustworthiness scores, and $T_y$, $T_n$ the sum of the trustworthiness scores of those  who answered `yes' and `no' respectively. The label is then determined by which of $T_y$ or $T_n$ is larger, in this case it is $T_y$, and the confidence score is $T_y/T$, in this case 0.78.

\begin{figure}[h!]
     \centering
     \begin{subfigure}[b]{0.5\textwidth}
         \centering
         \includegraphics[width=\textwidth]{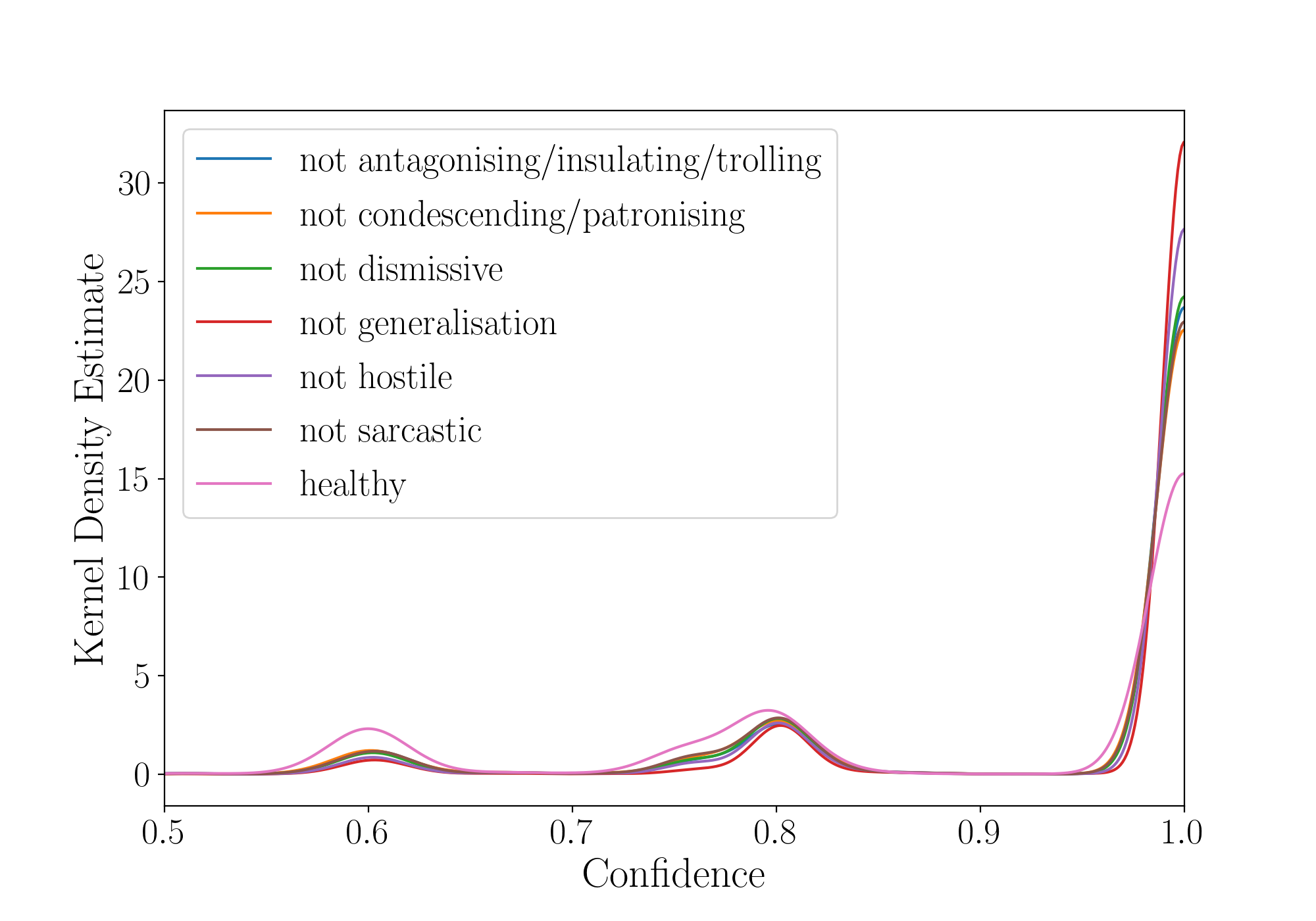}
         \caption{}
         \label{fig: densitiesno}
     \end{subfigure}
     
     \begin{subfigure}[b]{0.5\textwidth}
         \centering
         \includegraphics[width=\textwidth]{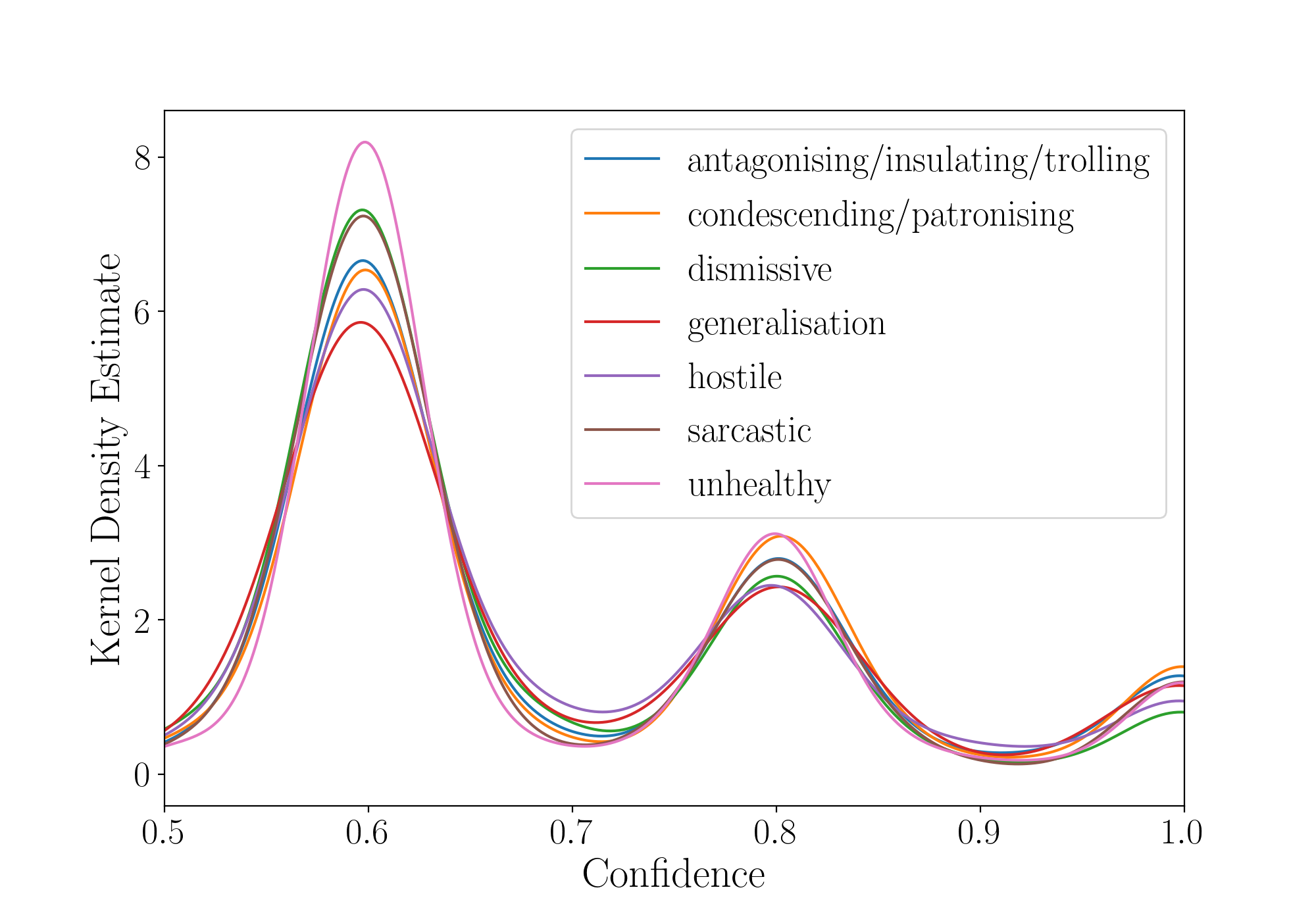}
         \caption{}
         \label{fig: densitiesyes}
     \end{subfigure}
     \caption{Density  estimation \cite{rosenblatt1956remarks}
     of  confidence scores for each attribute. Figure \ref{fig: densitiesno} shows confidence  scores for those comments labelled as 'no' for each unhealthy attribute,  while  Figure  \ref{fig: densitiesyes} represents those of  comments labelled 'yes'.}
     \label{fig: densities}
\end{figure}

 The proportion of comments that contain each attribute is shown in Table \ref{table: proportions} and the confidence distributions are shown in Figure \ref{fig: densities}.

\begin{table}[h!]
\begin{center}
\begin{tabular}{ | l | c |} 
\hline
 \textbf{Attribute} & \textbf{Proportion} \\ 
 \hline 
 Antagonistic/Insulting/Trolling & 4.7 \%\\  
 Condescending/Patronising & 5.5\%\\
 Dismissive & 3.1\% \\
 (Unfair) Generalisation & 2\%\\
 Hostile & 2.5\% \\
 Sarcastic & 4.3\%\\
 Unhealthy & 7.5\% \\
 \hline
\end{tabular}
\caption{Percentage of positive labels for each attribute.}
\label{table: proportions}
\end{center}
\end{table}

As the comments were sampled from the SFU Opinion and Comment Corpus dataset, the prevalence for each attribute is inevitably low. Despite the label imbalance, the dataset represents an important contribution to identification of this wider variety of subtle attributes, with thousands of positive examples for each. Our manual analysis during initial iterations of the annotation job indicated that these final proportions are roughly representative of the prevalence of these attributes in similar live contexts, such as North American  online newspaper comment sections. There are specific attributes, notably sarcasm, for which it can be possible to collate a corpus of self-labelled data, for example by scraping tweets with `\#sarcastic' from Twitter, or comments followed by `/s' on Reddit \cite{Khodak2018selfsarcasm}. In these specific circumstances, the avoidance of the need to crowdsource and pay for annotations can permit much larger and more balanced datasets. However, for all other attributes we consider, and in fora like the comment sections of news sites, relying on self-labelled data is not possible. For these attributes, crowdsourcing is the only feasible way to obtain high quality data, and as such we would expect proportions reflecting those observed in similar contexts. 

\begin{figure}
     \centering
     \begin{subfigure}[b]{0.5\textwidth}
         \centering
         \includegraphics[width=\textwidth]{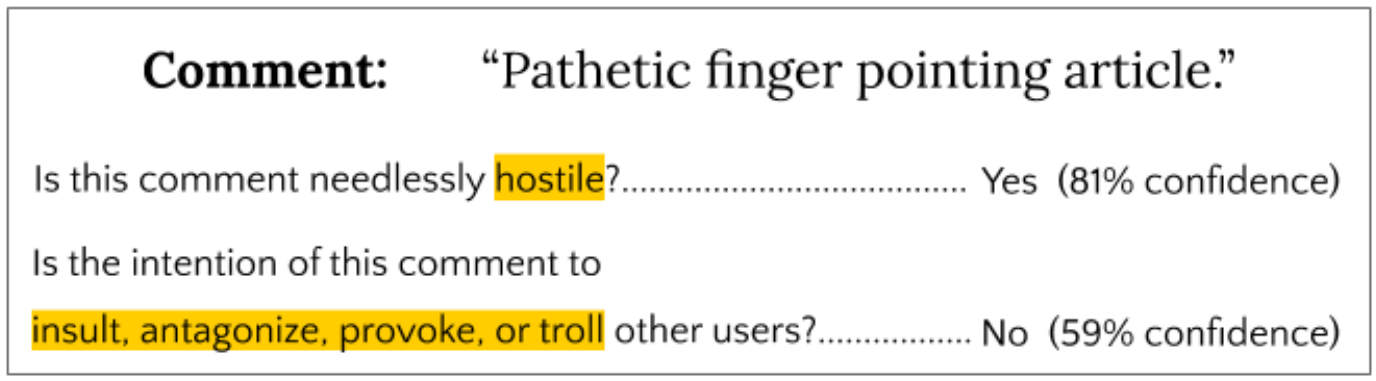}
         \caption{A distinction between a hostile comment and one which intends to insult, antagonize, provoke or troll other users.}
         \label{fig: subtle hostile}
     \end{subfigure}
     
     \vspace{0.3cm}
     
     \begin{subfigure}[b]{0.5\textwidth}
         \centering
         \includegraphics[width=\textwidth]{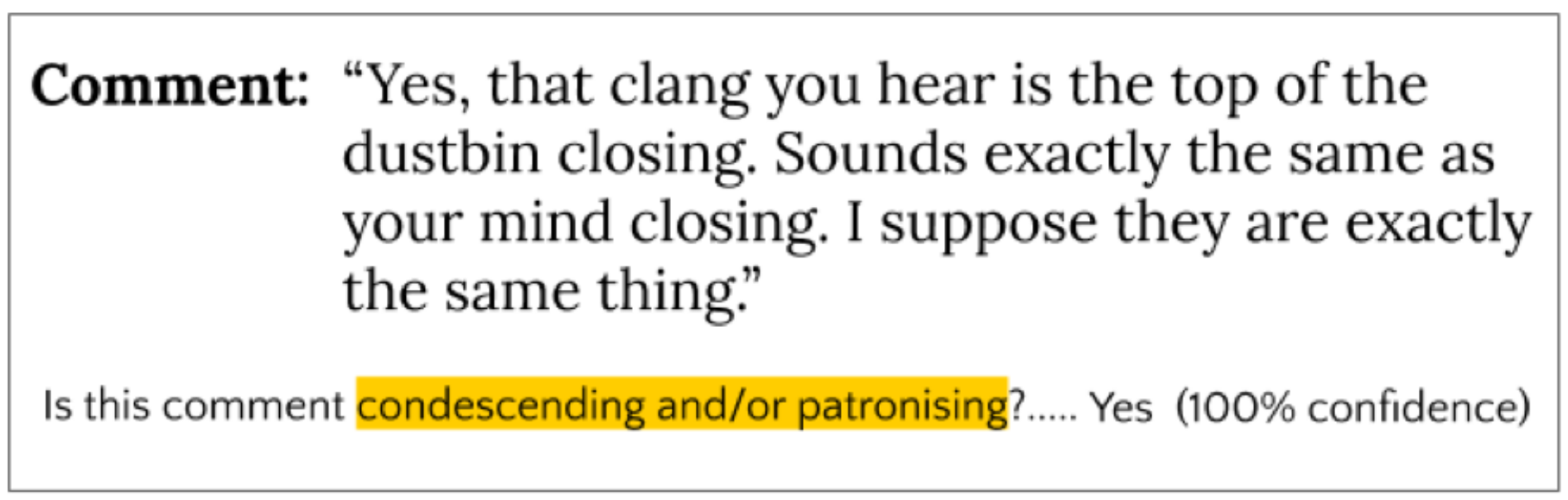}
         \caption{Subtle condescension}
         \label{fig: subtle cond}
     \end{subfigure}
     
     \vspace{0.3cm}
     
     \begin{subfigure}[b]{0.5\textwidth}
         \centering
         \includegraphics[width=\textwidth]{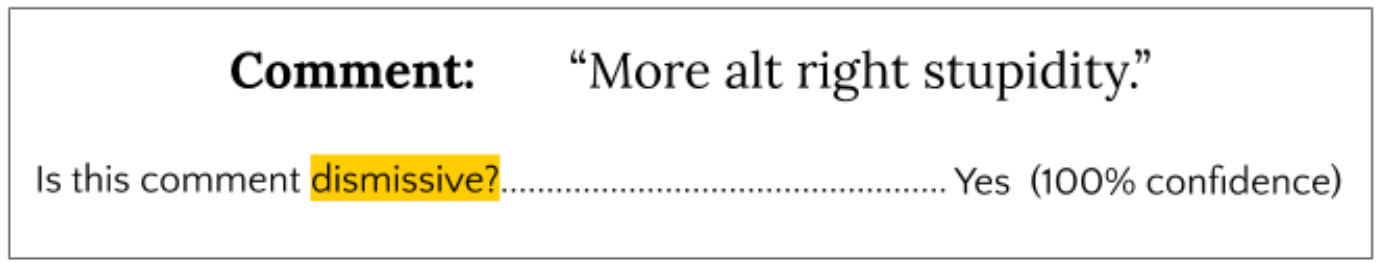}
         \caption{Implicit yet clear dismissiveness.}
         \label{fig: subtle  dis}
     \end{subfigure}
     
     \caption{Examples  of subtleties correctly picked  up by annotators,  with confidence scores shown in brackets alongside the resultant label.}
     \label{fig: subtle attributes}
\end{figure}

Inspection of random subsets of the new UCC dataset reveals that the data is generally of a high quality, and captures important nuances, accurately identifying these subtle attributes, both when they overlap (as is common), and also when they do not (see  Figure \ref{fig: subtle attributes} for examples).

\begin{figure}[!h]
    \hspace{0.1cm}
    \includegraphics[width = 0.48\textwidth]{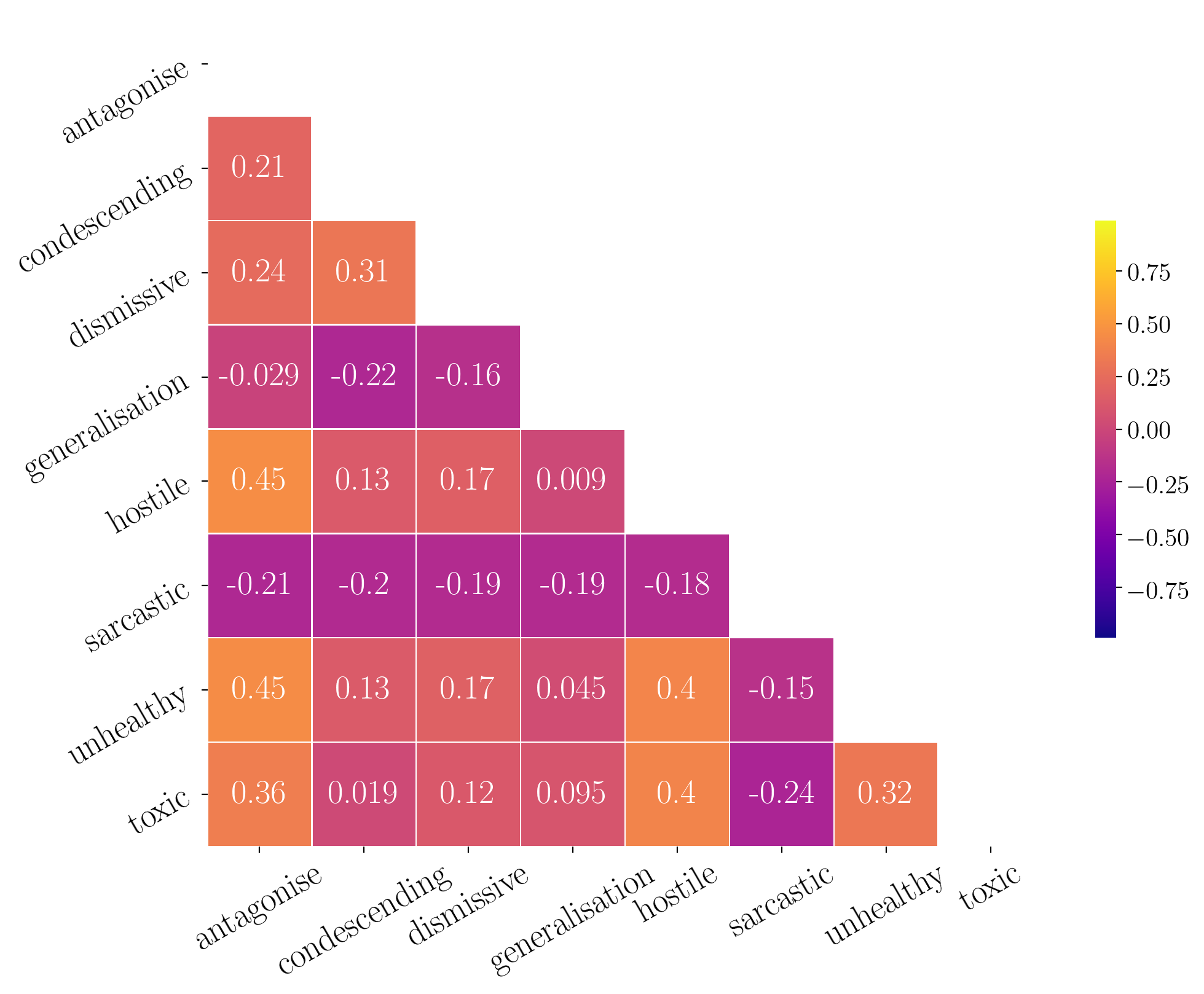}
    \caption{Inter-attribute correlations, including with `toxicity' as scored by Perspective API.}
    \label{fig:attribute correlations}
\end{figure}

Figure \ref{fig:attribute correlations} shows the correlations between attributes, calculated based on the pool of comments which are labelled as one or more of the six unhealthy attributes. The figure highlights two important facts. First, the relatively low correlation between most attributes indicates that the dataset succeeds in differentiating between these different types of subtle unhealthy attributes. As expected, there is significant correlation between antagonistic and hostile comments. There is some correlation between the often more subtle attributes like dismissiveness/condescension and antagonism, while these are less correlated with hostility. We  also  include correlations with the `toxicity' scores  produced by Jigsaw's Perspective API (\href{http://perspectiveapi.com}{perspectiveapi.com}), which again confirms that our attributes, in particular those other than antagonistic and hostile, capture something distinct from overt toxicity. A notable feature of Figure \ref{fig:attribute correlations} is the slightly negative correlations between sarcasm and other attributes, indicating that annotators generally did not associate sarcasm with other unhealthy attributes. Secondly, `unhealthy' correlates significantly with antagonism and hostility, but very little  with the other attributes, indicating a fairly broad general notion of healthy conversation on the part of the annotators, which mostly includes dismissive, condescending, sarcastic and generalising comments. 

Despite its generally high quality, the nature of the task and the annotation method entails some level of noise in the dataset. This noise is particularly difficult to quantify given the need to distinguish between different but reasonable interpretations of a comment, and simply incorrect annotations caused by a lack of understanding or care on the part of an annotator (for  example, one comment reading ``You are an ignorant $^*$sshole" was judged not to be needlessly hostile, an obvious error).

This highlights the difficulties of using traditional reliability metrics like Krippendorff's $\alpha$ for crowdsourced annotations on subjective tasks \cite{D'Arcey2019sarcasm}. Krippendorff's  $\alpha$ is a number between 0 and 1 intended to indicate the  extent to which annotators agree compared with what would have happened if they guessed randomly. The base assumption then is that all disagreement between annotators decreases reliability, which is not necessarily the case for subjective attributes \cite{salminen2018socialcomputing, swanson2014reliability}.

Despite the above caveat, we conduct analysis using Krippendorff's $\alpha$ (K-$\alpha$) for two reasons. Firstly, to allow for comparison  with other literature in the field, we report the K-$\alpha$ for judgements on each attribute in Table \ref{table: Kalpha}. They range from 0.31 - 0.39, which is comparable with other datasets labelling `similar' phenomenon, such as sarcasm (0.24-0.38) \cite{swanson2014reliability,justo2018detection,D'Arcey2019sarcasm}, and hate speech with sub-attributes from Figure Eight annotators (0.21) \cite{lazaridou2020bias}. The one exception is the set of judgements on whether a comment has a place in a healthy conversation, with a lower K-$\alpha$ of 0.26. Given that this is a more open-ended question, this is not necessarily surprising.

\begin{table}[h!]
\begin{center}
\begin{tabular}{ | l | c |} 
\hline
 \textbf{Attribute} & K-$\alpha$ \\ 
 \hline 
 Antagonistic/Insulting/Trolling & 0.39 \\  
 Condescending/Patronising & 0.36 \\
 Dismissive & 0.31 \\
 Generalisation & 0.35 \\
 Hostile & 0.36 \\
 Sarcastic & 0.34\\
 Unhealthy & 0.26\\
 \hline
\end{tabular}
\caption{Krippendorff's alpha by attribute.}
\label{table: Kalpha}
\end{center}
\end{table}

\begin{figure}[h!]
    \centering
    \includegraphics[width=0.5\textwidth]{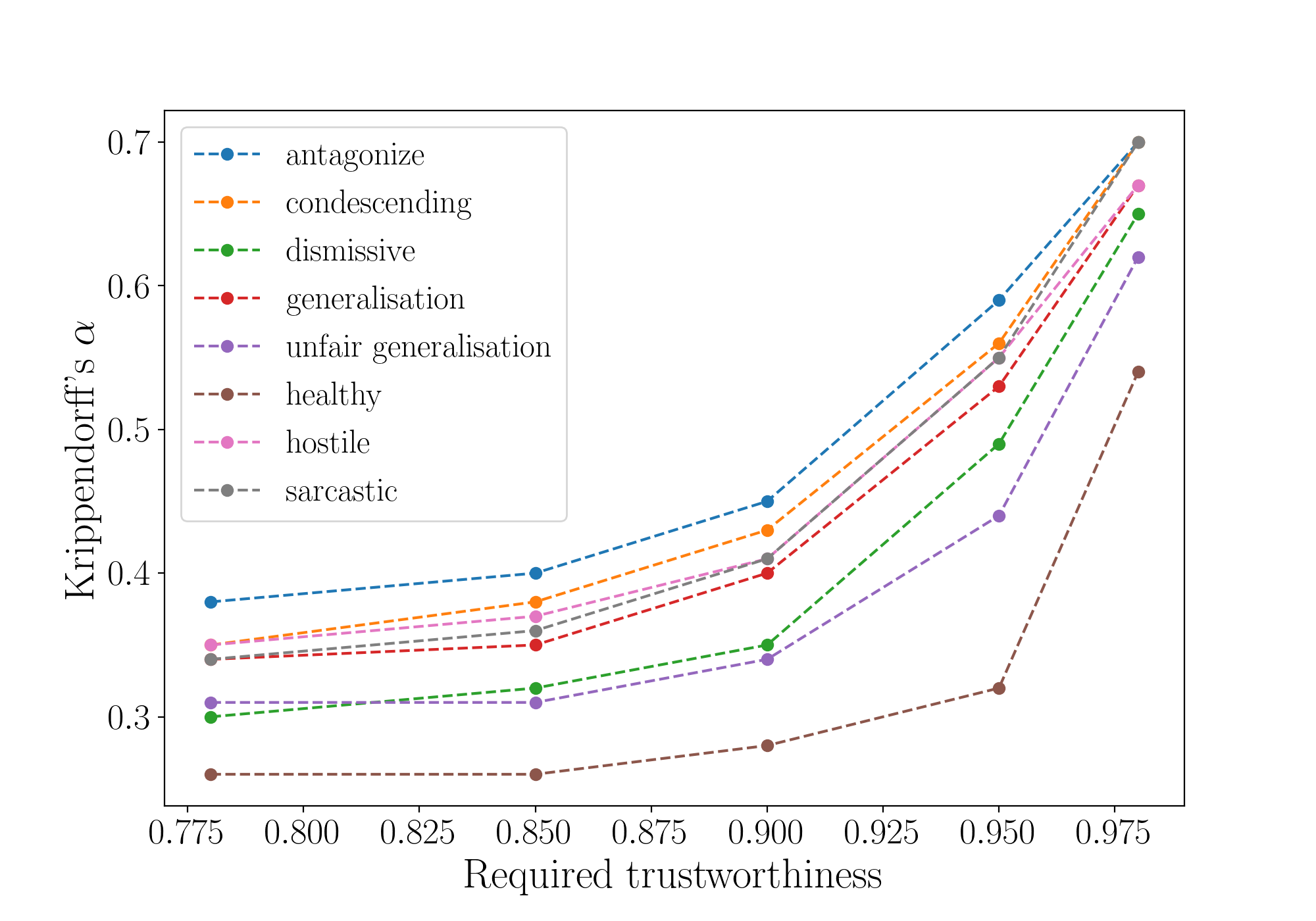}
    \caption{Krippendorff's $\alpha$ for various threshold levels of annotator trustworthiness.}
    \label{fig: K-alphas}
\end{figure}

Secondly, to the  extent that K-$\alpha$ is an important reliability metric for this form of data, it supports our use of `trustworthiness' scores when aggregating judgements on a given comment to decide labels and confidence scores. Specifically, as shown in Figure \ref{fig: K-alphas}, we see that as we increase the trustworthiness threshold for annotators whose judgements are included, the resulting K-$\alpha$ steadily increase. This provides some indication that our trustworthiness scores do capture the reliability of our annotators, and thus that their judgements ought to be weighted more highly in the final confidence in a comment's labels.

Also included in the UCC dataset are the individual annotations for each comment by all `trusted' annotators. Users of the data may therefore apply any alternative trustworthiness threshold, or use a preferred aggregation method to derive labels.
 
\section{Models and results}\label{sec: model}

\begin{figure}[h!]
    \centering
    \includegraphics[width=0.5\textwidth]{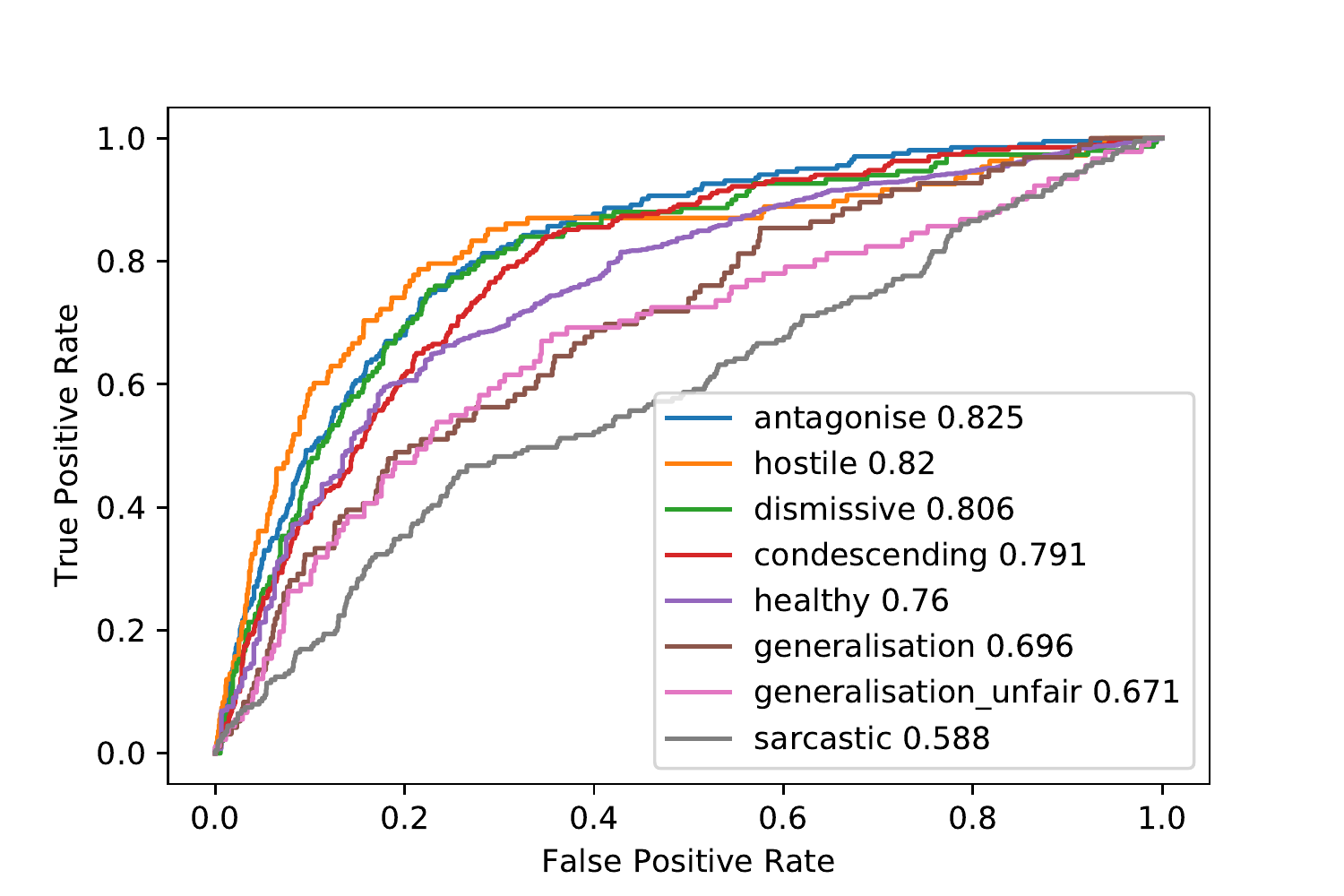}
    \caption{Receiver operating characteristic curves and  AUC for class each attribute.}
    \label{fig: auc}
\end{figure}

Use of a pre-trained BERT model \cite{Devlin2019BERTPO} and fine-tuning on this dataset produces classifiers with modest performance (Figure \ref{fig: auc}), compared to the state of the art for sequence classification. The best performing attributes, `hostile' and `antagonistic' are also those most similar to the types of attributes typically annotated in comment classification work. The other attributes seem to cluster together, with the `sarcastic' label particularly noteworthy for its low performance.

To give context to the model performance, we follow \cite{wulczyn2017ex} and compare our performance with human workers. For each comment, we hold out one annotator to act as our `human model' and use the aggregated score of the other annotators as the ground truth to compute the ROC AUC. To stabilize our results, this procedure is repeated five times and the average reported. We use the same test sets to compute the ROC AUC of the trained BERT model and average those scores as well. As we can see, for all attributes other than `sarcastic' the BERT model outperforms a randomly selected human annotator, indicating that it has sufficiently captured the semantic and syntactic structures for these attributes. For `sarcastic', the gap between the BERT model and human annotators indicates a rich area for studying whether model performance can be improved.

\begin{table}[h!]
\begin{center}
\begin{tabular}{ | l | c | c |} 
\hline
 \textbf{Attribute} & Human AUC & BERT AUC \\ 
 \hline 
 Antagonistic & 0.71 & 0.82 \\  
 Condescending & 0.72 & 0.78 \\
 Dismissive & 0.68 & 0.82 \\
 Generalisation & 0.73 & 0.74 \\
 Hostile & 0.76 & 0.84 \\
 Sarcastic & 0.72 & 0.64 \\
 Unhealthy & 0.62 & 0.69 \\
 \hline
\end{tabular}
\caption{Comparing Human and BERT performance}
\label{table: aucs}
\end{center}
\end{table}

\section{Potential Unintended Biases} \label{sec: biases}
One further challenge which comes with annotating more subtle unhealthy attributes is the potential to encode unintended societal biases and value judgements in models trained on this data. For example, sarcasm is often communicated by stating something which the author presumes to be so obviously untrue that it will be read as sarcastic. These presumptions reflect the author’s biases - or in the cases of comment annotation, labelling comments as sarcastic reflects the annotators beliefs of what is obviously untrue.

With the comment corpus being in English, and given the subtlety of the attributes, higher quality annotations were likely to be achieved by annotators with first-language proficiency in English. The best proxy for this available on the Figure Eight platform was to restrict the country of origin of our annotators to a limited subset of countries with a large English-speaking population (as either an official language or primary second language), in particular: the United States, the United Kingdom, South Africa, Sweden, New Zealand, Norway, Netherlands, Denmark, Canada, and Australia. Although our early iterations of this annotation job indicated a significant reduction in annotators failing test comments once this was enforced, this introduces a clear cultural and geographic bias. For example, the comment `Iran and Turkey are the BEST places to be a woman!', was scored as sarcastic with 72\% confidence by the annotators. Finding this comment sarcastic relies on an assumption by the annotators (of which the pool excludes residents of Iran and Turkey) that Iran and Turkey are clearly not the best places to be women. Our annotators were not selected as broadly representative across language, geography, culture, or other attributes and this assumption is not universal. While important research has begun to explore the composition of the global crowd workforce, it remains difficult to select for annotators representative of specific characteristics on crowd work platforms \cite{posch2018characterizing}. In the current version of the Appen platform, unless annotators are asked standalone questions on demographics, the only available details are the annotators' country and/or city (and even then, only for some annotators). Research and modelling based on this dataset, and similar datasets, requires the exercise of great care in mitigating biases produced by the underlying data collection. This potential selection bias is likely to be evident across the broader healthy/unhealthy categorisation along with each of the attributes. Prior research has found substantial disagreement on subtle attributes of speech both among individuals and across geographies \cite{salminen2018online}. 

Finally, the source of the comments and their manner of presentation could introduce bias into the dataset. The source data is solely from a Canadian online newspaper comment section and comments were presented in isolation to annotators, without the surrounding context of the news article and other comments. Annotators were also provided with the standard questionnaire (Appendix \ref{app: questionnaire}), which includes high level descriptions of the attributes that may not generalise across cultures. There is a substantial body of research demonstrating the potential impact of introducing biased datasets, and Vidgen \textit{et al.} \cite{vidgen2019challenges} note that public datasets in this area are prone to systematic bias and mislabelling, with inter-annotator agreement  typically low for complex multi-class tasks of this kind. These challenges are to be expected in a relatively new field which aims to improve on human baseline moderation for highly subjective characteristics of online discussion. At this early stage of research, we must be mindful of addressing these biases and cognisant that the manner in which this data is collected can have critical impacts on users in a production environment. It is important to note at this stage of the field in general, and with our understanding of this dataset in particular, that the UCC dataset is not designed to train models which are immediately available for automated moderation without human intervention in a live online setting. As the field develops further, initial use-cases may include less interventionist `nudges' or reminders of how a comment could be perceived by a reader to assist participants in discussions online.

\section{Conclusions and Further work}
 
We introduced a new corpus of labelled comments and a typology for some of the more subtle aspects of unhealthy online conversation. Our typology provides 6 sub-attributes of typically unhealthy contributions, and confidence scores for the labels. We described the process and challenges in creating such a dataset, and provided statistics to convey the scale of data. In particular, we note that although there is a substantial body of research on more extreme forms of negative contributions, such as toxicity, the subtler forms of unhealthy comments in our typology are often similarly prevalent online. Our analysis also shows that the sub-attributes are largely independent from overt toxicity, and mostly correlated with unhealthy contributions.

We also provide results from a modern baseline ML model (fine tuning BERT) and note that performance exceeds that of a crowd-worker. This suggests that further work could also be done to collect a larger corpus of annotations to improve the capacity to measure models in this domain. While this dataset provides a new contribution in gathering the 6 attributes under the umbrella of an `unhealthy' conversation, there also remains an open question as to how exhaustive this typology of unhealthy contributions is. Future research and annotation work could further refine the typology, amend the standard questionnaire, or apply it to forums which differ in cultural and geographic context.

Further work also includes exploring the unintended biases in the model and data. This dataset is well-placed to further explore early signs of conversations going awry \cite{zhang2018conversations}, while models based on the data could be explored to provide assistance to moderating online conversations.


\bibliographystyle{acl_natbib}
\bibliography{bibliography}

\newpage

\appendix

\section{Annotator Questionnaire} \label{app: questionnaire}

\noindent\textbf{Overview}

In this job, you will be asked to read a comment and to express an overall opinion about whether you think it has a place in a healthy conversation online. 

You will also be asked to identify whether it displays a range of characteristics that may lead to unhealthy conversations. These characteristics include: sarcasm, gross generalisations, hostility, aggression, dismissiveness, condescension and patronization. 

All of the comments you will see are real comments posted by users in online conversations. Most of them will have been posted in response to one or more comments made by others (which you are not given). However, the questions are designed in such a way that you should be able to answer them without seeing these other comments.

The data collected here will be used to help build tools which promote healthier conversations online.

\vspace{0.3cm}


\vspace{0.3cm}

\textbf{Note:}

\begin{itemize}
    \item Please bear in mind that the questions do not ask whether you agree or disagree with the substance of each comment. Do your best to ignore your own opinion on the substantive idea or claim made in the comment when answering the questions.
    \item Please be sure to read the full text of the comment before answering the questions. Sometimes the part of a comment which displays one or more of the attributes you will be asked about, appears close to the end of the comment.
\end{itemize}    

\vspace{0.3cm}


\vspace{0.3cm}

\begin{enumerate}
    \item Healthy Online Conversations:

What are the characteristics of a healthy conversation?

    \begin{itemize}
        \item Posts and comments are made in good faith
        \item Posts and comments are not overly hostile, and are not destructive
        \item The comments in the conversation generally invite engagement
        \item The conversation may include robust engagement and debate
        \item The conversation is generally focused on substance and ideas
    \end{itemize}

A healthy conversation does not necessarily require all posts and comments to be:

    \begin{itemize}
        \item friendly
        \item grammatically correct
        \item well constructed or well structured
        \item sanitized and free of any vulgarity
        \item intellectual or substantive
    \end{itemize}

With this in mind please answer the following question: Do you think this comment has a place in a healthy online conversation?
    \item A comment is sarcastic if it  uses irony in order to mock or convey contempt, or if its intended meaning is different from what is literally said. Sarcasm  can be used playfully, or harshly. Note: Not all humour (or nastiness) is sarcastic. 
    
    Is this comment sarcastic?
    \item Does this comment make a generalisation about a specific group of people?
    \item If yes, would a member of that group feel that the generalisation is unfair?
    \item Is this comment needlessly hostile?
    \item Is the intention of this comment to insult, antagonize, provoke, or troll other users?
    \item A comment with a condescending or patronising tone will generally assume an attitude of superiority, and imply that the other commenter(s) is ignorant, child-like, naive, or unintelligent. Such comments will usually imply that the other commenter shouldn't be taken seriously. 
    
    Is this comment condescending and/or patronising?
    \item A comment is dismissive if it rejects or ridicules another comment without good reason, or tries to push another commenter and their ideas out of the conversations. Note: A comment which expresses \textit{disagreement} is not necessarily dismissive. 
    
    Is this comment dismissive?   
\end{enumerate}

\end{document}